# TOWARDS A MORE USER-FRIENDLY CORRECTION

*Damien GENTHIAL, Jacques COURTIN, Jacques MÉNÉZO*

*Equipe TRILAN, LGI-IMAG Campus, BP 53, F-38041 Grenoble Cedex 9*
*E-Mail: Damien.Genthial@imag.fr*

## ABSTRACT

We first present our view of detection and correction of syntactic errors. We then introduce a new correction method, based on heuristic criteria used to decide which correction should be preferred. Weighting of these criteria leads to a flexible and parametrable system, which can adapt itself to the user. A partitioning of the trees based on linguistic criteria: agreement rules, rather than computational criteria is then necessary. We end by proposing extensions to lexical correction and to some syntactic errors. Our aim is an adaptable and user-friendly system capable of automatic correction for some applications.

## RÉSUMÉ

Nous présentons d'abord notre position par rapport à la détection et à la correction des erreurs syntaxiques. Nous introduisons ensuite une nouvelle méthode de correction qui s'appuie sur des critères heuristiques pour privilégier une correction plutôt qu'une autre. La pondération de ces critères permet d'obtenir un système souple et paramétrable, capable de s'adapter à l'utilisateur. Un découpage des arbres basé sur des critères linguistiques: les règles d'accord, plutôt que sur des critères informatiques est alors nécessaire. Nous terminons en proposant l'extension à la correction lexicale et à certaines erreurs syntaxiques. Notre objectif est un système adaptable, convivial et capable, pour certaines applications, de faire des corrections automatiques.

## 1. INTRODUCTION

Since 86, the TRILAN[1] team has taken an active interest in detection and correction of errors in French written texts. First centered on lexical errors (Courtin, 89), research work has since turned towards syntactic errors. Latter developments aim at building a complete system for detection and correction of errors (Courtin, 91), and even to define a more extensive Computer Aided Writing system (Genthial, 92).

In this kind of system, we have at our disposal a large number of modules, each with its own linguistic competence (morphology, phonetic, syntax). In this paper, we are interested in the correction process: the aim is to integrate at best the linguistic knowledge of each module in order to lead to a system capable of making automatic corrections (in a natural language man-machine interface), or almost automatic (in a computer aided writing system).

The paper is centered on agreement errors correction, specially frequent in French, but we hope to widen the technique to other kinds of errors.

## 2. DETECTION AND CORRECTION OF SYNTACTIC ERRORS

Any error which prevents the system from producing an interpretation (or more simply a parsing) for the input sentence is considered to be a syntactic error. These errors may be of very different kinds, but we can give two rough classes:

(a) errors due to the system: the input is correct but the linguistic coverage is insufficient;

(b) errors due to the user: the input is incorrect.

This classification, which can also be used for lexical errors, is far more relevant for the syntactic level because type (a) errors at this level are very frequent in free texts, such as newspaper articles for example. In order to avoid deadlocks due to these errors, one must build robust parsers, with wide coverage (Chanod, 91; Genthial, 90). We are going to concentrate here on type (b) errors.

We suppose the system has all the required competence and the deadlock is due to a misuse of the language by the user. We may then consider two ways to proceed:

---

[1]TRILAN : TRaitement Informatique de la LAngue Naturelle (Computational Treatment of Natural Language)

- either we relax constraints in order to obtain results, even incorrect, then we filter these results to find the origin of the error and finally correct it (Douglas, 92; Weischedel, 83);
- or we try to foresee the errors and we integrate in the grammar a way to express all possible types of errors, thus avoiding deadlocks of the parsing process (Goëser, 90).

We have chosen the first way because the richness of natural language makes it very difficult to describe all correct utterances. Therefore, it is in our opinion, impossible to enumerate exhaustively all possible errors, especially if we intend to verify texts read by automatic devices (scanners and characters recognition software).

The first method can be encountered for example in systems which aim to build a logico-semantic interpretation of the input sentence: in these systems, syntactic constraints are almost completely relaxed and parsing is based on semantic information (Granger, 83; Wan, 92).

We have therefore built a prototype (Courtin, 91) which can detect and correct agreement errors in number, gender and person, in simple French sentences. The most interesting feature of this prototype is not its coverage, which is limited, but the exhaustive design and implementation of all agreement rules of French grammar. It works as follows: we first make a morphological analysis of the input sentence, then we build all possible dependency structures for the sentence. Following the principle of relaxation of constraints, the process of building dependency structures does not take into account morphological variables, it uses only the lexical category of words. The resulting trees are then passed on to a checker which will attempt to verify the variables borne by the nodes, examining them by pairs, each pair composed of a governor and a dependant.

So to verify $les_{plu}$ $calcul_{sin}$ $scientifique_{sin}$[2] (*scientific computations*), we will first verify the pair ($calcul_{sin}$, $les_{plu}$) which is incorrect because of a disagreement in number between $calcul_{sin}$ and $les_{plu}$. We will then ask the user to choose between the two solutions : *les calculs* (plural) and *le calcul* (singular). In order to generate these solutions, we use a morphological generator which is of course based on the same data as the morphological parser mentioned above.

The user's choice is then introduced in the tree and the verification process resumes. If the user chose the plural, we will have an error again with ($calculs_{plu}$, $scientifique_{sin}$) leading to a new, obviously useless, question to the user.

This traversing of trees using pairs has proved useful to design agreement rules, but is clearly not adapted to a user-friendly correction. Moreover, it does not take into account the context of the incorrect pair. We therefore propose first the use of correction strategies and then a new way of traversing the trees which are to be verified.

## 3. USING CORRECTION HEURISTICS

By definition of the concept of agreement error, every such error always gives two lexical units which may be corrected. The choice of the unit to be corrected is left to the user but we think that in most cases the proper correction can be chosen automatically. Actually, when a human being rereads a text, even if he is not the author, he very rarely hesitates between the two possible corrections of an agreement error. One can always say that a human reader understands the written text but we can also imagine simple heuristics (i.e. machine computable), which could allow correction without hesitation.

### 3.1. Heuristics

For examples of such heuristics, we could have (Véronis, 88, quoted in Genthial, 92):
a) number of errors in a group: $le_{sin}$ $vélos_{plu}$ $est_{sin}$ $redevenu_{sin}$ *à la mode* will be corrected in the singular *le vélo est redevenu à la mode* (only one word corrected), rather than the plural *les vélos sont redevenus à la mode* (three words corrected with, moreover, an alteration of the meaning, very hard to detect with simple techniques);
b) it is better to correct in a way that does not modify the phonetics of the sentence: $Les_{mas,fem}$ $chiens_{mas}$ $dressées_{fem}$... will be corrected in the masculine *Les chiens dressés...* rather than the feminine *Les chiennes dressées...* We find here again the idea, often used at the lexical level, that incorrect written utterances follow the phonetics of the correct form.
c) writer laziness: a writer sometimes omits an *s* where one is necessary, but rarely adds

---

[2] As it is not easy to find good examples of complex agreement errors in english, we use French examples but we make explicit the variables causing trouble : here the number with *sin* for singular and *plu* for plural.

one where it is not: $les_{plu}$ $enfant_{sin}$... is thus corrected as $les_{plu}$ $enfants_{plu}$....

d) one can give priority to the head of the phrase (underlined): $les_{plu}$ $petits_{plu}$ $\underline{enfant}_{sin}$ $qui$ $ont_{plu}$... becomes singular $le$ $petit$ $enfant$ $qui$ $a$... The idea here is that the writer takes more care of the main word of a phrase than of the others.

We could also find other criteria, by studying corpora or by interviewing professionals such as teachers of French or journalists.

These heuristics are of course open to criticism, the main argument against them being that they are no longer valid with the use of text editors because cutting and pasting of portions of text may introduce errors which would not have been made in linear writing.

Moreover, they are often conflicting: consider for example the sentence $j'aime$ $les_{plu}$ $calcul_{sin}$ $scientifique_{sin}$ which includes an agreement error in number. The (a) criterion leads to correct $les_{plu}$ in $le_{sin}$ because 2 words among 3 are singular. The $s$ not being pronounced at the end of French words, the (b) criterion leads to correct plural $les$ $calculs$ $scientifiques$, without phonetic alteration. The (c) criterion imposes the plural and the (d) criterion the singular of $calcul$, which is the governor.

### 3.2. Weightings

Despite everything, we can hope to obtain automatic corrections thanks to the use of more than one criterion and if we are able to weight the various criteria in order to compute a confidence factor for each correction.

Consider for example, for the above criteria, that the confidence factor is computed with the following formulae:

a) $K_a * \dfrac{1 + \#\_of\_correct\_words}{1 + \#\_of\_corrected\_words}$

b) $\dfrac{K_b}{1 + \#\_of\_phonetic\_alterations}$

c) $K_c$

d) $K_d$

where the $K_i$ are weights assigned to each criterion. We will take $K_a = 2$, $K_b = 2$, $K_c = 2$ and $K_d = 1$.

If we apply these weightings to $les_{plu}$ $calcul_{sin}$ $scientifique_{sin}$, we get Table 1.

*Table 1 : Example of weightings*

|          | (a) | (b) | (c) | (d) |
|----------|-----|-----|-----|-----|
| singular | 3   | 1   | 0   | 1   |
| plural   | 4/3 | 2   | 2   | 0   |

A null value fits a case where the confidence factor can not be evaluated: thus for the (c) criterion we can only correct in plural and for the (d) criterion, on this example, singular is imposed by the governor.

If we sum the factors of each row, the correction *j'aime les calculs scientifiques* (plural) wins by 5,33 (51,6%) against 5 (48,4%) for *j'aime le calcul scientifique* (singular). It is true that in this case, the weakness of the difference makes it advisable to ask the user to choose his correction, but we can decide to use a threshold T such that, if the absolute value of the difference between the two confidence factors (0.3 on the example) is above T, correction will be automatically done for the solution with the higher confidence factor.

### 3.3. Adaptability

One of our hypotheses is that the value and thus the weight of a correction criterion depends on a given user or at least on a given class of users (scientists who master the language but not the keyboard, children or foreigners learning the language, secretaries who master both keyboard and language but are inattentive,…).

Consequently, we want to build a system where the criterion weights are not fixed, but may be dynamically updated by means of a simple learning mechanism. Initially, weights are either arbitrarily chosen, or chosen following the assignment of the user to a particular class, and the automatic correction threshold is set very high. With that configuration, most errors lead to a consultation of the user and his answer is used to increase the weight of those criteria which would have selected the proper answer and to weaken the weight of the others.

In the above example, if the user forces the singular, the system will increase the weight of the (a) and (d) criteria and weaken the weight of (b) and (c).

In the same way, the threshold will decrease each time the weights are modified until it reaches a lower limit, arbitrarily fixed or chosen by the user.

However, the implementation of these correction criteria in a verification-correction system for agreement errors assumes that the minimal unit of correction, which was a pair (governor, dependant) in the prototype described in §2, must be redefined in order to render possible the evaluation of the confidence factor for each correction proposal.

## 4. A NEW CORRECTION METHOD

Consider for example the sentence:
*les$_{plu}$ jeunes$_{plu}$ cycliste$_{sin}$ que j'$_{sin}$ai$_{sin}$ rencontré$_{sin}$ montaient$_{plu}$ à bon$_{mas}$ allure$_{fem}$*[3].

It contains an agreement error in gender between *bon$_{mas}$* and *allure$_{fem}$*, and two agreement errors in number: one in the nominal phrase: *les$_{plu}$ jeunes$_{plu}$ cycliste$_{sin}$* and the other between the subject *cycliste$_{sin}$* and the verb *montaient$_{plu}$*. If we choose to correct this sentence by forcing the plural, we introduce a new error between the past participle *rencontré$_{sin}$*, and its object complement *cyclistes*, which has became plural. The associated dependency tree is shown in Fig. 1.

*Fig. 1: Example of a dependency tree*

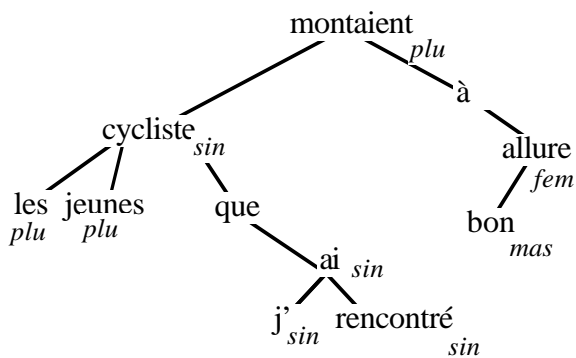

The agreement rules which apply are then:
- agreement between determiners, adjectives and noun inside a nominal phrase;
- agreement between the past participle of the relative clause *rencontré* and its object *cycliste* because it is placed before;
- agreement between the subject and the verb;
- agreement between the subject and the auxiliary *ai* in the relative clause.

Reading these rules suggests dividing the verification-correction problem according to agreement dependency existing between the nodes of the tree. We then apply the following method:

1) Partitioning of the tree in three sub-trees, each one connected, but not necessarily disconnected two by two. There must exist a dependency between the variables (gender, number, person,…) of the nodes of a sub-tree but no dependency between the sub-trees themselves:

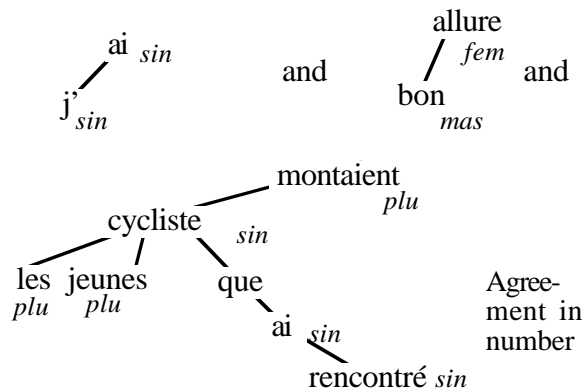

2) Checking of agreement rules for each sub-tree obtained: here we exploit the previous work by verifying only those rules which have decided that a sub-tree was actually one. We verify by the classical method of tree traversing with unification of the values of variables. We then eliminate the group *j'ai*, which is correct.

3) If at least one error is detected in a group, we must attempt to correct it by using the heuristics defined above. For *bon$_{mas}$ allure$_{fem}$*, we will correct in the feminine *bonne allure* because *allure* has no masculine.

3.1) However, it is interesting to divide complex groups into more simple ones, always according to the agreement rule involved. In the example, we will divide the first group, which includes the relative clause, into the three sub-trees of Fig. 2.

*Fig. 2: Partitioning of the tree*

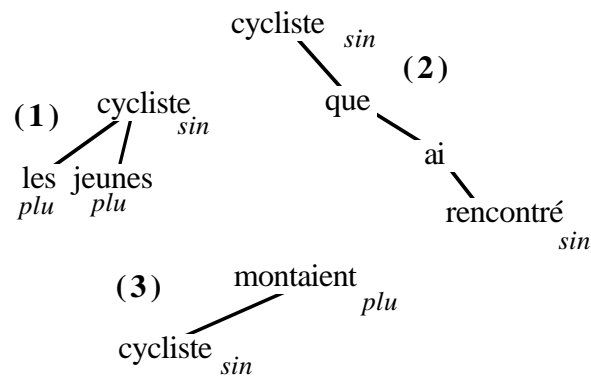

Such a partitioning is interesting because the agreement error in number, detected on the whole group does not appear in all the sub-groups. If we attempt to correct separately each sub-group (with the criteria and the weights defined above) we obtain Table 2.

---

[3]Something like: *the young cyclist I have met were climbing at good speed.*

*Table 2: Confidence factors by sub-groups*

|     | (a) sin | (a) plu | (b) sin | (b) plu | (c) sin | (c) plu | (d) sin | (d) plu |
|-----|---------|---------|---------|---------|---------|---------|---------|---------|
| (1) | 4/3     | 3       | 1       | 2       | 0       | 2       | 1       | 0       |
| (2) | 6       | 2/3     | 2       | 2       | 0       | 2       | 1       | 0       |
| (3) | 2       | 2       | 2       | 2       | 0       | 2       | 0       | 1       |

When summing the confidence factors of the various criteria, we obtain Table 3.

*Table 3: Sums of the confidence factors*

|     | singular       | plural          |
|-----|----------------|-----------------|
| (1) | 3,33 (32,25%)  | 7 (67,75%)      |
| (2) | 9 (65,85%)     | 4,66 (34,15%)   |
| (3) | 4 (40%)        | 6 (60%)         |

If the threshold T is small enough ($< 2$), we can consider *les jeunes cyclistes (plural)* as the good correction for the first sub-group, the second sub-group is correct and the plural corrects the third. But these results leave an error on the whole group.

3.2) So we must evaluate the whole group correction by using the results of each sub-group. Here again, we can exploit various criteria of evaluation:
- simple majority: we choose the most frequently selected correction in the sub-groups. Plural wins by 2 to 1. We could also weight each group according to the number of words or to statistical criteria on errors: agreement errors on past participles used with the auxiliary *avoir* (have) are especially frequent in French, due to the complexity of the rules involved; so the weight of the second sub-group would be lowered.
- proportional majority: we sum the confidence factors of all sub-groups for each possible correction. This leads to correction in the plural (17,66) rather than the singular (16,33). We can here again use a threshold below which the conclusion is not considered reliable.
- weighted proportional majority which uses the percentages and so is a mixture of the two previous ones: we sum the percentage of each sub-group. Plural wins by 161,9 against 138,1 for the singular. Comparing with the previous method, we weaken the importance of the second sub-group which, being correct, has a big difference between the two confidence factors.

In the example, the plural wins, but when it is not possible to automatically choose the good correction, the choice is left to the user. It is then very interesting to exploit the partitioning of the tree to ask a very relevant question to the user: the intersection of the three sub-trees is the word *cycliste*, so we can question the user as follows:

In the sentence:
*les jeunes cycliste que j'ai rencontré montaient à bonne allure.*
Did you want to say *un cycliste (singular)* or *des cyclistes (plural)* ?

According to the answer, the whole sentence is corrected and possibly the weights and the threshold are updated.

## 5. EXTENSIONS

With these correction methods, the organisation of the correction system is less deterministic. By this, we mean that it is easier to modify its behaviour by updating the weights or the thresholds or by adding new verification rules. This flexibility should make it easier to process syntactic ambiguities due to the relaxation of constraints during the parsing process. For example the sentence: *la maison de l'oncle que nous avons vu(e)* (*the house the uncle we have seen*) produces two trees in French if we do not consider agreement rules in gender, but produces only one if we do, depending on the gender of the past participle *vu(e)*. If it is feminine then we have seen the house, if it is masculine then we have seen the uncle. A correction system must then, whenever one of the two trees is correct, apply correction rules to both of them in order to detect a possible error. This implies that we imagine a traversing method of all the trees of the same sentence at the same time. We are at present working on this question.

The techniques presented above and the correction module which will result are designed for a complete correction system where many modules cooperate in a client/server architecture. We shall then extend the use of weights to the lexical level, for which we have implemented 3 correction techniques: similarity keys, phonetics and morphology (Courtin, 91; Genthial, 92). Each of these techniques proposes, for an incorrect word, a list of correction hypotheses which must be sorted in decreasing likelihood order so that we give the user only the more likely ones. We will weighting each technique and the values of weights will follow dynamically the types of errors of a given user, thus allowing an alternative implementation of the architecture proposed in (Courtin, 89).

Some lexical errors can only be detected at superior levels (syntactic even semantic) like *I to not want* for *I do not want* or *the doc barks* for *the dog barks*. These errors, named hidden errors (Letellier, 93), lead to a blocking of the syntactic parsing. Here again, the use of prediction mechanisms (syntactic parser or statistical model based on Markov chains), coupled with a weighting of the proposed solutions must allow some automatic corrections below a given threshold.

Finally, we think it is possible to implement a system making completely automatic corrections. The §4 example is described in the framework of a computer aided writing system, able to deal with free texts for which it is very hard and even impossible to produce a complete semantico-pragmatic interpretation. On the other hand, if we try to build a robust man-machine interface, then we can hope for a completely automatic correction because:
- in this type of applications, the lexicon is very limited, so the number of corrections for a lexical error will be small;
- lexical ambiguities will also be less numerous and therefore the number of trees produced will be lower;
- we can use, to resolve syntactic ambiguities or to refine the above criteria, some semantic or pragmatic information which can be well defined because of the restricted domain.

## 6. CONCLUSION

The TRILAN team has at its disposal the basic tools necessary in order to build such a system: we have the morphological tools (analysis and generation), the phonetic tools (graphic ↔ phonetic transducers) and the syntactic tools (dependency structure builder and agreement rules). We have started a project of "lingware" engineering to make all these tools work together in a client/server architecture. We will integrate in all the linguistic servers the possibility of weighting their results each time they give multiple solutions. The detection and correction system itself will be basically a controller, managing the answers of the various servers and the variations of weights and thresholds, in order to make the system fit to a particular user. Our aim is to obtain a general and flexible system which could fit into various applications (text processing, man machine interface, computer aided translation).